\documentclass[letterpaper, 10 pt, conference]{ieeeconf}  

\IEEEoverridecommandlockouts                              
        
\usepackage{cite}
\usepackage{amsmath,amssymb,amsfonts}

\usepackage{algorithmic}
\usepackage{arydshln}
\usepackage{graphicx}
\usepackage{textcomp}
\usepackage{xcolor}
\usepackage{multirow}
\def\BibTeX{{\rm B\kern-.05em{\sc i\kern-.025em b}\kern-.08em
    T\kern-.1667em\lower.7ex\hbox{E}\kern-.125emX}}
\usepackage{mathtools}

\usepackage{siunitx}
\usepackage{url}
\usepackage{overpic}
\usepackage{pgf}
\usepackage{microtype}
\usepackage{booktabs}
\usepackage{todonotes}
\usepackage{color, colortbl}  

\usepackage{hyperref}
\hypersetup{colorlinks,
linkcolor=[rgb]{0.5,0.,0.},
citecolor=[rgb]{0.000,0.427,0.173},
urlcolor=black
}

\usepackage{subcaption}
\def\secref#1{Sec.~\ref{#1}}
\def\figref#1{Fig.~\ref{#1}}
\def\tabref#1{Tab.~\ref{#1}}
\def\eqref#1{Eq.~(\ref{#1})}

\newcolumntype{C}[1]{>{\centering\let\newline\\\arraybackslash\hspace{0pt}}m{#1}}
\newcolumntype{R}[1]{>{\raggedleft\let\newline\\\arraybackslash\hspace{0pt}}m{#1}}

\title{\LARGE \bf SiLVR: Scalable Lidar-Visual Reconstruction with Neural Radiance Fields for Robotic Inspection}

\author{Yifu Tao$^{1}$, Yash Bhalgat$^{2}$, Lanke Frank Tarimo Fu$^{1}$, Matias Mattamala$^{1}$, Nived Chebrolu$^{1}$, and Maurice Fallon$^{1}$%
}

\begin{document}

\twocolumn[{%
\renewcommand\twocolumn[1][]{#1}%
\maketitle
\begin{center}
    \vspace{-25pt}
    \centering
	\includegraphics[width=1\textwidth]{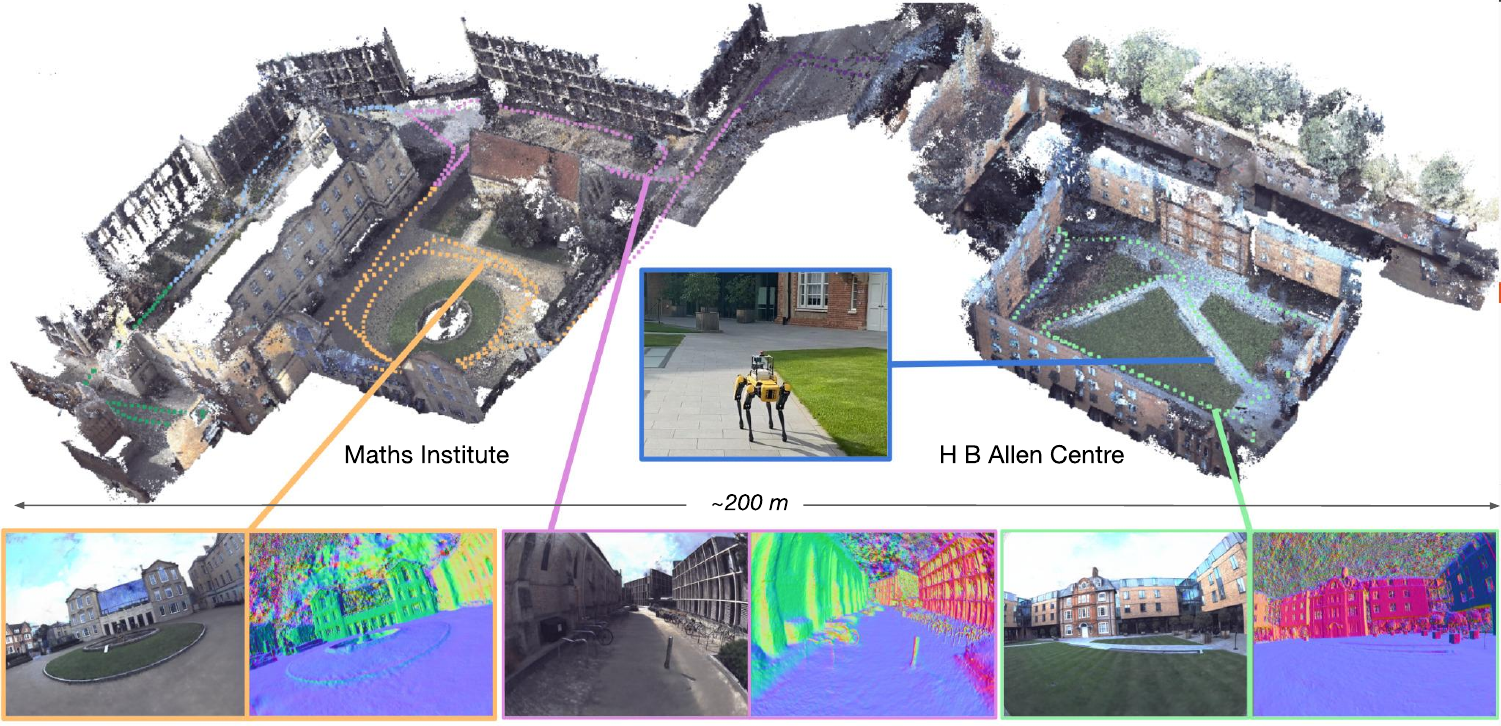}
	\captionof{figure}{Large-scale reconstruction consisting of 8 submaps of Maths Institute and H B Allen Centre in Oxford. The bottom row shows the novel views synthesised from the model and surface normals at three different locations. The trajectory of each submap is visualised in a different colour. 
	}
    \label{fig:header}
\end{center}%
}]

\begin{abstract}
We present a neural-field-based large-scale reconstruction system that fuses lidar and vision data to generate high-quality reconstructions that are geometrically accurate and capture photo-realistic textures. This system adapts the state-of-the-art neural radiance field (NeRF) representation to also incorporate lidar data which adds strong geometric constraints on the depth and surface normals. We exploit the trajectory from a real-time lidar SLAM system to bootstrap a Structure-from-Motion (SfM) procedure to both significantly reduce the computation time and to provide metric scale which is crucial for lidar depth loss. 
We use submapping to scale the system to large-scale environments captured over long trajectories. We demonstrate the reconstruction system with data from a multi-camera, lidar sensor suite onboard a legged robot, hand-held while scanning building scenes for 600 metres, and onboard an aerial robot  surveying a multi-storey mock disaster site-building. Website: \url{https://ori-drs.github.io/projects/silvr/}
\end{abstract}

\begingroup
  \renewcommand\thefootnote{}\footnote{$^1$Oxford Robotics Inst., Dept. of Eng. Science, Univ. of Oxford, UK. \{\href{mailto:yifu@robots.ox.ac.uk}{\nolinkurl{yifu}}, \href{mailto:fu@robots.ox.ac.uk}{\nolinkurl{fu}}, \href{mailto:matias@robots.ox.ac.uk}{\nolinkurl{matias}}, \href{mailto:nived@robots.ox.ac.uk}{\nolinkurl{nived}}, \href{mailto:mfallon@robots.ox.ac.uk}{\nolinkurl{mfallon}}\}\texttt{@robots.ox.ac.uk}\\
  $^2$Visual Geometry Group, Dept. of Eng. Science, Univ. of Oxford, UK. \href{mailto:yashsb@robots.ox.ac.uk}{\nolinkurl{yashsb}} \}\texttt{@robots.ox.ac.uk}.\\
  This project has been partly funded by the Horizon Europe project Digiforest (101070405). Maurice Fallon is supported by a Royal Society University Research Fellowship. For the purpose of open access, the authors have applied a Creative Commons Attribution (CC BY) license to any Accepted Manuscript version arising.}
  \addtocounter{footnote}{-1}%
\endgroup

\section{INTRODUCTION}
Dense 3D reconstruction is a task that underpins a range of robotic applications such as industrial inspection and autonomous navigation. Common sensors used for reconstruction include cameras and lidar. Camera-based reconstruction systems use techniques including Structure-from-Motion (SfM) and Multi-View Stereo (MVS) to produce dense textured reconstructions~\cite{schoenberger2016sfm}. However, these systems rely on good lighting conditions as well as having observed the many view constraints. They also struggle with textureless areas. Lidar provides accurate geometric information at long range --- as it directly measures distances to surfaces. This makes it desirable for large-scale outdoor environments, but the sensor measurement is usually much sparser than a camera image. It also does not capture colour which is important for applications such as virtual reality and 3D asset generation.


Classical reconstruction systems have used point clouds, occupancy maps, and sign-distance fields (SDF) as their internal 3D representation. Recently, neural radiance field (NeRF) \cite{mildenhall2021nerf} has gained popularity for visual reconstruction. With differentiable rendering, this approach optimises a continuous 3D representation by minimising the difference between the rendered image and reference camera images. It can achieve state-of-the-art novel view synthesis quality.

As with traditional vision-based reconstruction methods, NeRF struggles to estimate accurate geometry in locations where there is limited multi-view input and little texture. Autonomous systems commonly encounter these situations - for example, reconstructing a wall with uniform colour when traveling directly towards it. This problem could be addressed by using lidar sensing to give accurate geometric measurements in these textureless objects. In addition, the use of lidar data can mitigate the need to capture multiple camera views. It is impractical for an inspection robot (e.g. The Spot quadruped on an industrial facility) to execute object-centric trajectories just to improve the visual reconstruction. This motivates the development of a reconstruction system that fuses these sensors. 

In this work, we present a NeRF-based reconstruction system that integrates both lidar and visual information to generate accurate, textured 3D reconstruction which also provides photo-realistic novel view synthesis. Our method builds upon NeRF implementation \cite{nerfstudio} utilising hash encoding \cite{mueller2022instant} that takes minutes to achieve photo-realistic rendering. This is extended with geometric constraints from lidar to improve reconstruction quality. The use of lidar enables depth measurements even from featureless areas, which cannot be obtained from SfM \cite{deng2022depth}. Surface normal can also be computed from a lidar scan, which is more robust than learning-based priors \cite{Yu2022MonoSDF} which can suffer from input data distribution shift in real-world deployment.

This system is demonstrated using data from a perception sensor suite which contains three wide field-of-view cameras and a 3D lidar to enable robots to reconstruct in all directions. We take advantage of a lidar-inertial odometry and SLAM mapping system \cite{wisth2023vilens} as part of the pipeline. Experiments are presented using a drone, a legged robot, and a handheld device in industrial and urban environments.

In summary, our main contributions are:
\begin{itemize}
    \item A dense textured 3D reconstruction system that achieves accurate geometry that's on-par with lidar, and photorealistic novel view synthesis
    \item Integration with a lidar SLAM system, so that the NeRF is trained with both depth and surface normal obtained from lidar data, and metric-scale trajectory with a reduction in computation time by 50\% compared with up-to-scale offline Structure-from-Motion method commonly used in literature.
    \item A sub-mapping system that scales to large outdoor environments --- with trajectories over 600 metres
    \item Evaluation of the system on real-world large-scale outdoor datasets captured from multiple robotic platforms.
\end{itemize}

\section{RELATED WORKS}
\label{sec:related_works}

\subsection{Large-scale 3D Reconstruction}
Lidar is the dominant sensor for accurate 3D reconstruction of large-scale environments \cite{behley2018rss, lin2022r3live}, thanks to its accurate, long-range measurements. 
Volumetric lidar mapping relies on high-frequency odometry estimates from scan-matching and IMU measurements \cite{zhang2014loam, zhao2021super, xu2022fast, wisth2023vilens}. Yet, lidar-based systems may yield partial reconstructions, especially when robots explore scenes and have limited field-of-view sensors. Visual cues can be used to densify lidar mapping \cite{tao2022pdc}.

Alternatively, SfM systems such as COLMAP~\cite{schoenberger2016sfm} can generate large-scale textured reconstructions from images by first estimating camera poses using sparse feature points in a joint bundle adjustment process, followed by refinement with multi-view stereo algorithms~\cite{furukawa2010dense}. However, they face challenges in low-texture areas, repetitive patterns, and feature matching across views can be problematic with changing lighting or non-Lambertian materials. Urban Radiance Fields \cite{rematas2022urban} proposes using lidar sweeps along with RGB images to optimise a neural radiance field model that can be used for 3D surface extraction. Our work shares this approach, fusing both sensor modalities to generate precise geometry, overcoming limitations of vision-only approaches in low-texture areas, and being much denser than lidar-only reconstructions.

A common strategy to extend dense reconstructions to large-scale areas is through submapping \cite{bosse2003atlas, ho2018virtual,reijgwart2020voxgraph, wang2022strategies, tancik2022block}. These approaches partition the scene into individual local submaps which can incorporate the effect of loop closure corrections while still producing a consistent global map while achieving significantly lower runtime. Our work also adopts the submapping approach and partitions large-scale scenes into local maps (approximately 50x50m) using a globally-consistent lidar SLAM trajectory. This increases the representation capability and improves reconstruction, especially for thin objects.




\subsection{Neural Field Representation}
Neural Radiance Fields (NeRF) \cite{mildenhall2021nerf} proposed using a multilayer perceptron (MLP) to represent a continuous radiance field with differentiable volume rendering to reconstruct novel views. It implicitly induces multi-view consistency with geometric priors in the learned encodings.
NeRF and its many variants used frequency encoding \cite{vaswani2017attention} to encode spatial coordinates, but these suffer from long training times, typically a few hours per scene. Alternative explicit representations of radiance fields, including dense voxel-grids with trainable per-vertex features \cite{fridovich2022plenoxels,mueller2022instant} and more recently 3D Gaussians \cite{kerbl3Dgaussians} are shown to accelerate the training, at the cost of being more memory intensive.
Octree or sparse-grid structures \cite{yu2021plenoctrees,mueller2022instant} can reduce memory usage by pruning grid-features in empty space. 
Our work is built upon Nerfacto \cite{nerfstudio} which incorporates the main features from other NeRF works \cite{mueller2022instant,barron2022mipnerf360,martinbrualla2020nerfw} that have been found to work well with real data.


While NeRFs excel at high-quality view synthesis, obtaining a 3D surface from these representations remains challenging, mainly due to the flexible volumetric representation being under-constrained by the limited multi-view inputs. One approach to improve the reconstruction is to impose depth regularisation \cite{deng2022depth, rematas2022urban} or surface normal regularisation \cite{Yu2022MonoSDF}.
Another approach is to impose surface priors on the volumetric field and use 
representations such as Signed Distance Field (SDF) \cite{neuralangelo,yariv2021volume} to enforce a surface reconstruction output, although the novel view synthesis quality might be compromised \cite{wang2021neus} with this approach. 
Our method is built upon a volume density representation which is extended with depth \cite{deng2022depth} and surface normal \cite{Yu2022MonoSDF} regularisations from lidar measurements instead of using SfM \cite{deng2022depth} or learnt priors \cite{Yu2022MonoSDF}. In particular, it can significantly improve the reconstruction quality in texture-less areas with smooth surface.

Neural field representations have been used for lidar-base mapping \cite{zhong2023icra,deng2023nerfloam}, showing promise in generating more complete and compact reconstructions than traditional methods. While these works also build upon implicit map representation, they do not use visual data for building the map. Our system uses visual information and multi-view geometry constraints, therefore can reconstruct regions outside lidar field-of view.

\section{METHOD}
\label{sec:Method}

\begin{figure}[h]
	\centering
	\includegraphics[width=\columnwidth]{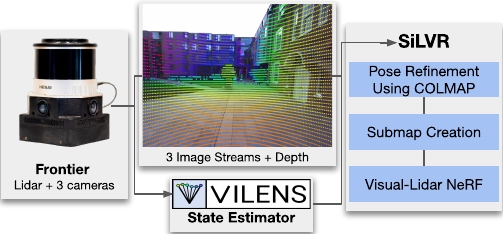}
	\caption{System Overview: Frontier, our custom perception unit, has three fisheye colour cameras with an IMU and a 3D lidar. Our online state estimator's trajectory is refined with COLMAP and partitioned into submaps. The camera image, lidar depth, and normal image are used to train a NeRF to get the final 3D reconstruction.}
	\label{fig:pipeline}
\end{figure}

We present a system for large-scale 3D reconstruction based on a NeRF representation tailored for robotic inspection tasks. We use a custom-designed sensor payload with a 3D lidar sensor, three fisheye cameras, and an IMU, which is suitable for use on various robot platforms. In \secref{sec:method_geometry_constraints}, we present our approach to fusing both lidar and vision information during the optimization phase to ensure high-quality reconstruction. Furthermore, we employ a submapping strategy to scale the approach to large areas described in~\secref{sec:method_submapping}. An overview of the system is presented in~\figref{fig:pipeline}.

\subsection{NeRF-based Scene Representation}
Our work builds on the differentiable volume rendering framework used for novel view synthesis \cite{mildenhall2021nerf,lombardi2019neural, sitzmann2019deepvoxels, sitzmann2019scene}.
To render a novel view from a NeRF given a viewpoint, we cast rays from the camera origin along the viewing direction for each pixel $u$ in the image plane, and render the pixel-colour by integrating over points sampled along the ray. This volume rendering integral is approximated using quadrature rule \cite{max1995optical,kajiya1984ray} as $\hat{c}_u = \sum_{i=0}^{N} w_i c_{i}$, where
\begin{equation} \label{eq:rendering}
w_i =   \exp{\left(-\sum_{j=1}^{i-1} \delta_j \sigma_j\right)} (1-\exp{(-\delta_i\sigma_i)}).
\end{equation}
Here, $\sigma_i$ and $c_i$ are the predicted density and color for the sampled 3D points and $\hat{c}_u$ is the rendered pixel color.

Our implementation is built on top of the Nerfacto method from Nerfstudio \cite{nerfstudio}. Nerfacto's rendering quality is comparable to state-of-the-art methods such as MipNeRF-360 \cite{barron2022mipnerf360} while achieving a substantial acceleration in reconstruction speed since it also incorporates efficient hash encoding from Instant-NGP \cite{mueller2022instant}. We also use scene contraction proposed in \cite{barron2022mipnerf360} to improve memory efficiency and represent scenes with high-resolution content near the input camera locations. The contraction function non-linearly maps any point in space into a cube of side length 2, and represents the scene within this contracted space. Since there is large variation in exposure and lighting conditions, we use a per-frame appearance encoding for each image, similar to \cite{rematas2022urban,martinbrualla2020nerfw}.

\subsection{Geometric Constraints from Lidar Measurements}
\label{sec:method_geometry_constraints}

3D reconstruction with NeRF becomes challenging when the surface has uniform texture and limited multi-view constraints. Lidar measurements are complementary as it can provide accurate measurements in such scenarios. In our work, we incorporate the lidar measurements in the NeRF optimization. Specifically, we impose a lidar-based depth regularisation by adding a depth loss \cite{deng2022depth} defined as the KL-Divergence between a normal distribution around the lidar depth-measurement $\mathbf{D}$ and the rendered ray distribution $h(t)$ from the NeRF model:
\begin{equation}
 \label{eq:depth}
\mathcal{L}_{\text {Depth }}=\sum_{\mathbf{r} \in \mathcal{R}} \mathrm{KL}[\mathcal{N}(\mathbf{D}, \hat{\sigma}) \| h(t)]    
\end{equation}

We also run a semantic segmentation network \cite{wu2019detectron2} to obtain a sky mask, and minimise the weights of these rays similar to \cite{rematas2022urban}.  

While the depth loss improves 3D reconstruction, we found that the surface contains wavy artifacts in regions where it is expected to be smooth (see Fig.\ \ref{fig:surface_normal}). To mitigate this, we compute the surface normal as the negative gradient of the NeRF's density field, and impose a further surface normal regularisation loss, inspired by \cite{Yu2022MonoSDF}:
\begin{equation}
	 \label{eq:normal}
    \mathcal{L}_{\text {normal }}=\sum_{\mathbf{r} \in \mathcal{R}}\|\hat{N}(\mathbf{r})-\bar{N}(\mathbf{r})\|_1+\left\|1-\hat{N}(\mathbf{r})^{\top} \bar{N}(\mathbf{r})\right\|_1
\end{equation}


\begin{figure*}[t]
	\centering
	\includegraphics[width=1.98\columnwidth]{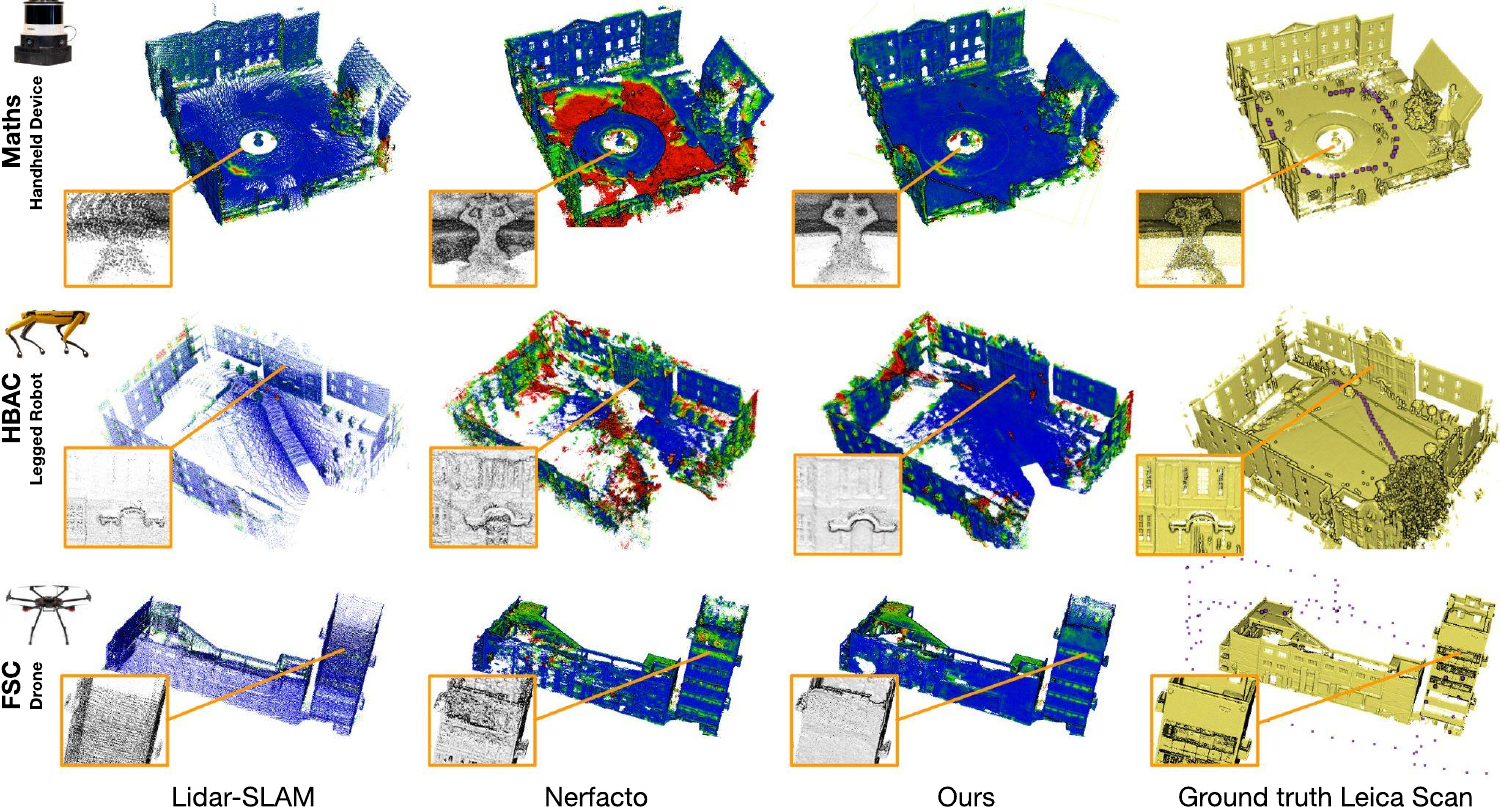}
	\caption{Comparison of reconstruction quality of Lidar-SLAM, Nerfacto (vision-only) and our approach. Reconstructions are coloured with point-to-point distance to the ground truth with increasing error from blue (0m) to red (1m). The trajectory is shown in purple and overlaid on the ground truth scan captured using a Leica BLK360. The zoomed-in views show the difference in reconstruction quality. Overall, our approach is more complete w.r.t lidar-only reconstruction, and geometrically more consistent w.r.t vision-only reconstruction.}
	\label{fig:recon_compare}
\end{figure*}

\subsection{Bootstrapping Camera Poses from SLAM with scale}
\label{sec:method_pose}
Obtaining accurate camera poses is crucial as the pose accuracy directly impacts the fidelity of the reconstructed model. A popular approach used in most NeRF works is to obtain camera poses using offline Structure-from-Motion methods such as COLMAP \cite{schoenberger2016sfm}. However, COLMAP has the following limitations: (1) long computation time, especially for large image collections collected over a long trajectory, and (2) inability to register all frames into one global map when there is limited visual overlap between the images. These issues limit its application in building a large-scale globally consistent map for robotic applications.

In our work, we use our lidar-inertial odometry and SLAM system VILENS \cite{wisth2023vilens}. While VILENS achieves state-of-the-art results for online motion tracking, we found that the camera poses obtained are less accurate than those of COLMAP. This results in \textit{blurring} artifacts in the renderings of the NeRF model. Several works \cite{tancik2022block,azinovic2022neural} use noisy pose inputs and then jointly refine the poses in parallel with the NeRF optimization to generate better results. In our experiments with a collection of large number of images, we found that while this pose-refinement approach sometimes leads to slightly better PSNR (peak signal-to-noise ratio), the resulting rendering is still less sharp compared to using the COLMAP estimated poses, and the training process is usually unstable.

To overcome the above limitations, we use the SLAM poses as a pose prior and refine the trajectory using COLMAP. Specifically, we replace the COLMAP \textit{mapper} with \textit{point triangulator} which reads prior poses. 
This method has the advantage of being faster, as it converts the incremental Structure-from-Motion into a global bundle adjustment problem, and more importantly, results in COLMAP being able to merge all available images in a single map. 
For a mission spanning over 20 minutes, our COLMAP-with-prior pipeline achieves similar rendering quality, while only taking half the computation time compared to a fully offline COLMAP run. The computation time is similar to the robot's mission duration, making it more applicable for robotic applications. 

After COLMAP computation, we rescale the trajectory using Sim(3) Umeyama alignment with lidar-slam trajectory, so that the final trajectory is metric. This is essential to use lidar measurements in \ref{sec:method_geometry_constraints}, since lidar depth is also metric.


\subsection{Scaling NeRF with Submapping}
\label{sec:method_submapping}

Training a NeRF for large-scale scenes is challenging as a single NeRF model has limited representation power and hardware constraints such as RAM usage when loading thousands of images. 
We adopt a sub-mapping approach, and partition the COLMAP-refined SLAM trajectory into clusters using Spectral Clustering \cite{von2007tutorial}. Different from \cite{tancik2022block}, our representation is based on Instant-NGP \cite{mueller2022instant} which is orders of magnitude faster than a Multilayer Perception. The submaps are trained in their local coordinate frames, and the final reconstruction are \textit{Sim(3)} transformed to the world coordinate frame using their metric pose from \ref{sec:method_pose}.

To generate 3D reconstructions from NeRF, we sample the rays used to train each model and render the colour and depth to generate a 3D point. 
When evaluating the 3D reconstruction, we found that submaps contain artifacts, especially around their boundary. One cause of the artifacts is that the boundary regions are observed sparsely with limited view constraints. To tackle this, we identify regions with low surface density and remove them when merging submap clouds to get the final reconstruction.

\section{EXPERIMENTAL RESULTS}
\label{sec:experiments}
\subsection{Hardware and Datasets}
We evaluate our methods on a custom perception unit called Frontier, a multi-camera lidar inertial device. It includes three 1.6 MP colour\footnote{Raw Bayered images are processed using \url{https://github.com/leggedrobotics/raw_image_pipeline}} fisheye Alphasense cameras on 3 sides of the device, with an IMU from Sevensense Robotics AG and a Hesai Pandar QT64 lidar.  We used the Frontier to collect datasets on multiple platforms: a legged robot (Boston Dynamics Spot), a drone (DJI M600), and a hand-held device. Spot and handheld Frontier datasets were taken at the H B Allen Centre (HBAC) and Mathematical Institute (Maths Inst.) in Oxford. The DJI M600 drone was operated at the Fire Service College (FSC). In the FSC dataset, we only use the rectified front camera image as the drone propellers were visible in the left and right camera images. 

We use VILENS \cite{wisth2023vilens}, a lidar-inertial SLAM system online to provide a globally consistent trajectory and motion-corrected lidar measurements. The online SLAM trajectory is further optimised by COLMAP offline, which improves the visual reconstruction quality, as shown later in \tabref{tab:pose_ablation}. The lidar point clouds are projected as a depth image. The surface normal is computed from the lidar range image, and also projected as a normal image. We calibrate our cameras-to-IMU extrinsics with Kalibr \cite{furgale2013unified} and cameras-to-lidar with \cite{fu2023extrinsics}. When running COLMAP, we further optimise the intrinsics from Kalibr. For training the NeRF model, we used an Nvidia RTX 3080 Ti and one iterations takes 4096 rays.

\subsection{Evaluation Metrics}
To evaluate the geometry of the reconstruction, we report \textit{Accuracy} and \textit{Completeness} following the conventions of the DTU dataset \cite{aanaes2016large_dtu}. Accuracy is measured as the distance from the reconstruction to the reference 3D model (ground truth) and encapsulates the reconstruction quality. Completeness is the distance from the point-wise reference to the reconstruction and shows how much of the surface is captured by the reconstruction. 
For the ground truth, we use the centimetre-accurate point cloud captured by a survey-grade Leica BLK360 laser scanner and register the lidar point clouds using transformation obtained from ICP, which is also used to register NeRF reconstructions.

We also evaluate the visual quality by reporting the Peak Signal-to-Noise Ratio (PSNR) and Structural Similarity Index (SSIM) \cite{wang2004image}. Note that our images have various exposure times, which lower the test PSNR even if the reconstructed image is photorealistic. 

\subsection{Evaluation of the 3D Reconstruction}
We perform a quantitative evaluation of our method on real-world datasets captured by different robotic platforms. We compare the point cloud reconstructions generated with the following configurations:
\begin{enumerate}
    \item Lidar-SLAM: lidar point cloud registered with SLAM poses refined by COLMAP
    \item Nerfacto \cite{nerfstudio}: baseline method using only the images
    \item SiLVR: Our method using photometric loss, depth loss, and surface normal loss
\end{enumerate}

We sample the training rays and generate 10 million points from each submap for Nerfacto and SiLVR. 
For Nerfacto, we excluded the points belonging to the sky by computing a sky segmentation mask with further manual cropping. 
For lidar-slam, we only include points within the camera field-of-view. This omits points pointing backwards that may be occluded by the operator.

\begin{table}[h]
	\caption{\small{ Evaluation of 3D Reconstruction Quality}}
   	\setlength{\tabcolsep}{3pt} 
	\centering
	\begin{tabular}{ l c c c c c}
		\toprule
		Method& Accuracy$\downarrow$ & Completeness$\downarrow$ &  \multicolumn{2}{c}{PSNR$\uparrow$} & SSIM$\uparrow$ 
		\\
            &(m)&(m)&train&test&test
            \\
            \hline
		\addlinespace
		\textbf{Maths Quad} 
		\\
		\hline
		\addlinespace
		lidar-SLAM	&\textbf{0.06}&0.15 & / & / & /
		\\
		Nerfacto mono &1.38&0.33&\textbf{31.3}&17.7&0.65
            \\
            Nerfacto 3-cam &0.76 &0.31& 28.0&\textbf{21.1}& \textbf{0.72}
		\\
		Ours mono &0.08&0.12&30.1&17.5&0.61
            \\
            Ours 3-cam &0.08 &\textbf{0.11} &27.3&20.5&0.71
		\\
		\hline
		\addlinespace
		\textbf{Oxford HBAC}
		\\
		\hline
		\addlinespace
		lidar-SLAM &\textbf{0.05}& 0.25& /&/&/
		\\
		Nerfacto mono&0.49&5.40 &\textbf{32.6}&19.5&0.65
            \\
            Nerfacto 3-cam&0.28&0.40&29.8&20.6&\textbf{0.74}
		\\
		Ours mono& 0.30&4.60&31.0&\textbf{21.2}&\textbf{0.74}
            \\
            Ours 3-cam &0.09&\textbf{0.18}&28.8&19.7&\textbf{0.74}
		\\
  		\hline
		\addlinespace
            \textbf{FSC} 
		\\
		\hline
		\addlinespace
		lidar-SLAM		&\textbf{0.08} &\textbf{0.08}&/&/&/
            \\
		Nerfacto mono &0.14&0.11 &\textbf{28.8}&\textbf{19.1}&\textbf{0.76}
		\\
		Ours mono&0.11 &0.09&27.7&\textbf{19.1}&0.75
		\\
		\bottomrule
		\addlinespace
	\end{tabular}
	\footnotesize{
 NeRF models are trained for 10000 iterations for 5 minutes; When computing accuracy and completeness, we crop regions where
 there is a change or no overlap between the ground truth and lidar scans.
} 
	\label{tab:reconstruction_eval}
\end{table}

We summarise the quantitative results in \tabref{tab:reconstruction_eval}, and show 3D reconstructions in \figref{fig:recon_compare}. Compared to Nerfacto, our method incorporates lidar measurements and has significantly better geometry which is shown in terms of accuracy and completeness. Nerfacto struggles when reconstructing the uniformly-coloured ground in Maths Inst., and the quad area in HBAC where the robot only walked forwards. 
Compared to lidar-SLAM, our method generally achieves more complete reconstruction since it uses dense visual information, while the accuracy (8-11cm) is nearly on-par with lidar-slam (6-8cm) and much better than Nerfacto (14-76cm). 

\begin{figure}[h]
	\centering
	\includegraphics[width=0.98\columnwidth]{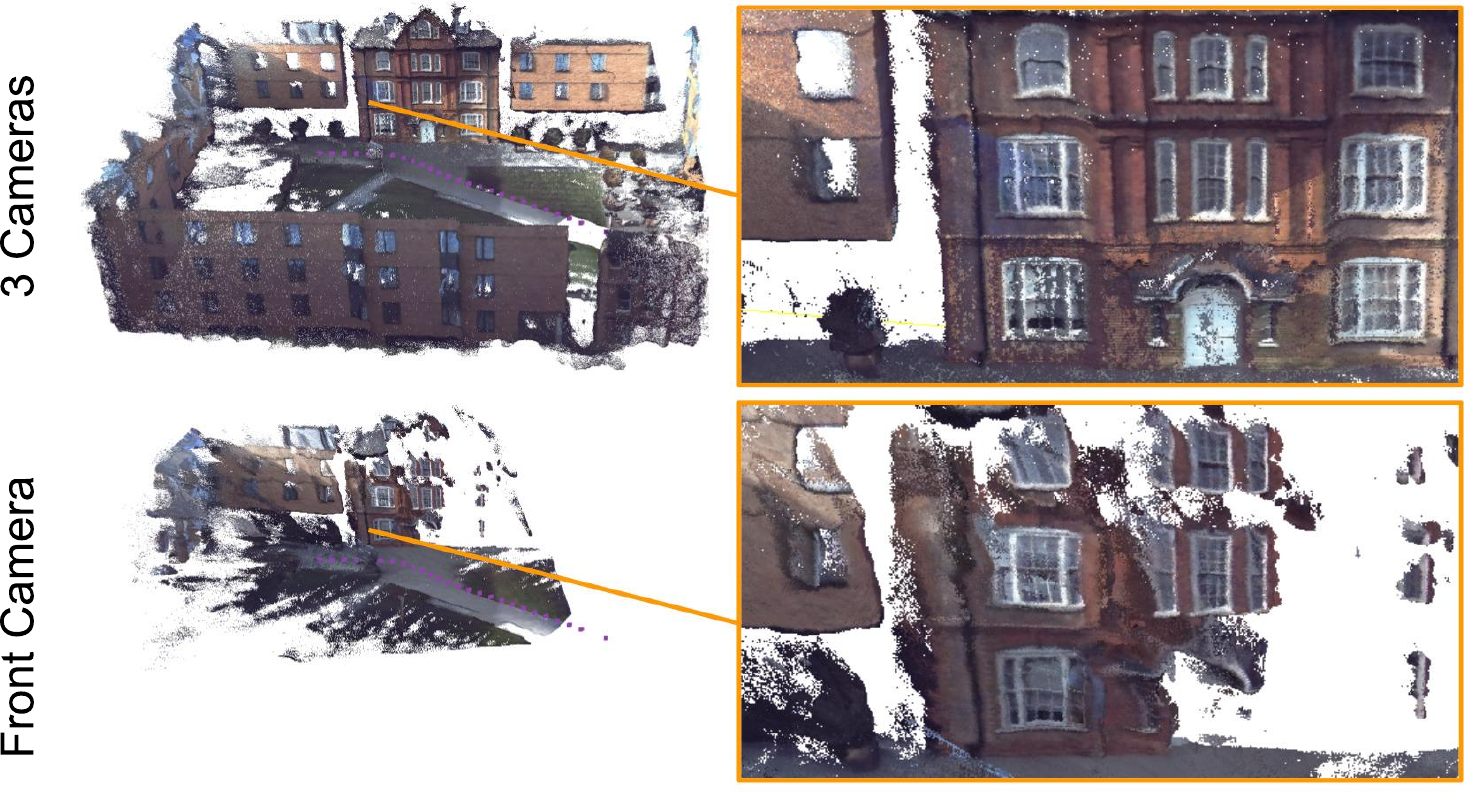}
	\caption{Comparison of reconstruction of HBAC building using the front camera only vs. using all the three cameras. The three-camera setup generates more complete and accurate reconstructions compared to using only a single front-facing camera. The multi-camera setting is important in robotic applications where it would be infeasible to actively scan the entire scene to obtain strong viewpoint constraints.}
	\label{fig:mono_3cam}
\end{figure}

The advantage of our multi-camera sensor setup is demonstrated qualitatively in \figref{fig:mono_3cam}. Compared to the three-camera setup, using only the front-facing camera leads not only to an incomplete reconstruction but also worse geometry. Visual reconstruction with photometric loss is biased towards generating a good rendering only at the input viewing angle. The reconstruction using the front-only camera in \figref{fig:mono_3cam} is trained with images viewing a shallow angle of the scene, and results in a poor geometric reconstruction when rendered from an unseen angle.

\subsection{Effect of Lidar Surface Normal Loss}

\begin{figure}[t]
	\centering
 
	\includegraphics[width=0.99\columnwidth]{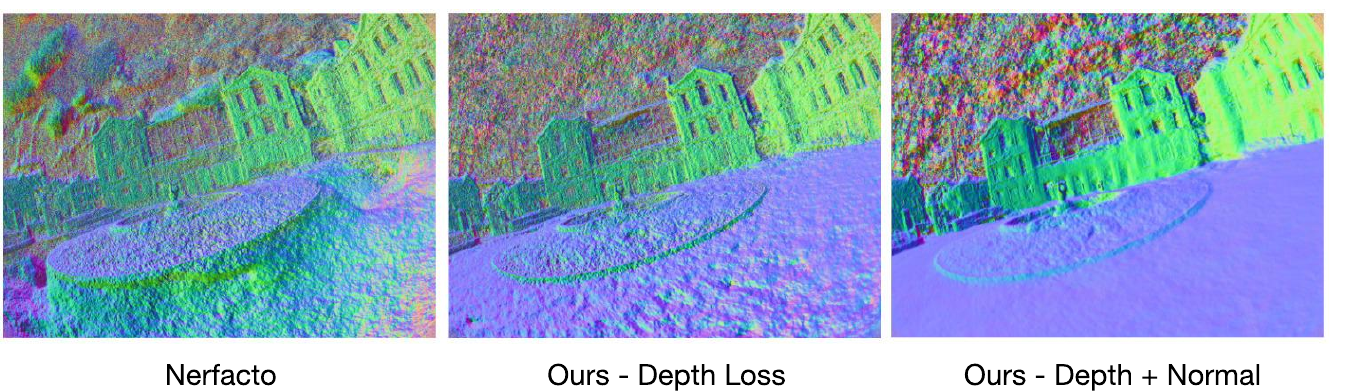}
	\caption{Comparison of surface normal renderings of the Maths Institute. Incorporating normal constraints in addition to depth from lidar improves the smoothness of the reconstruction. Right: The smooth reconstruction of the ground portion highlights this improvement.}    
 \vspace{-9pt}
	\label{fig:surface_normal}
\end{figure}

While the depth loss~\eqref{eq:depth} provides geometric constraints, we observe that the resulting 3D reconstruction is not necessarily smooth for flat surfaces. The results in \figref{fig:surface_normal} demonstrate how the surface normal loss~\eqref{eq:normal} further constrains the reconstruction geometry and improves the reconstruction quality. Nerfacto fails to estimate the depth of the ground due to its uniform texture. Using the lidar depth loss ensures ground reconstruction is at the right height, however, the surface is still not smooth. The surface normal loss furthers smoothens the surface and results in a higher-quality reconstruction.

\begin{table}[h]
	\caption{\small{Ablation: Effect of Bootstrapping SLAM Poses}}
  	\setlength{\tabcolsep}{3pt} 
	\centering
 	\begin{tabular}{ l c c R{1.0cm} r r r r}
		\toprule
		Method & Features & Prior& Traj. Regis  & \multicolumn{2}{c}{PSNR$\uparrow$} &SSIM$\uparrow$   & Time 
		\\
            &&& tered & Train&Test&Test
		\\
            &&&(\%)&&&&(s)
            \\
		\hline
		\addlinespace

            VILENS &/&/&100.0&23.0&17.4&0.64&Online
            \\
            NeRF refined &/&/&100.0&23.2&17.9&0.65&Online
            \\
            \addlinespace

            \multirow{4}{1cm}{COLMAP Sequential} & 1024&   &57.6&25.9&19.1&0.71&3299.2
            \\
            &1024 &\checkmark &100.0&26.2&\textbf{20.6}&\textbf{0.74}&1729.9
            \\
            &8192 &&94.0&26.1&19.8&0.72&7850.0
            \\
            &8192 &\checkmark &100.0&26.2&20.4&0.73&4448.4
            \\
		\addlinespace
            \multirow{4}{1cm}{COLMAP VocabTree} & 1024&   & 54.7&26.2&19.0&0.71&4444.8
            \\
            &1024 &\checkmark & 100.0&26.3&20.4&0.73& 1052.5
            \\
            &8192  &&94.8&\textbf{26.6}&19.9&0.72&37067.5
            \\
            &8192 &\checkmark &100.0&26.3&20.4&\textbf{0.74}&11015.3
            \\
		\bottomrule
		\addlinespace
	\end{tabular}
	\footnotesize{Results evaluated on HBAC-Maths dataset with 3254 images and duration of 1270s. Models trained for 4000 iterations. 
 PSNR and SSIM are evaluated after masking out the sky.}
	\label{tab:pose_ablation}
\end{table}

\subsection{Effect of Bootstrapping SLAM Poses}
We compare the performance of different strategies for computing poses: online SLAM poses, SLAM poses with NeRF pose refinement~\cite{nerfstudio}, SLAM poses with COLMAP~\cite{schoenberger2016sfm} in different configurations, and COLMAP without any prior poses. For COLMAP, we tested different numbers of features extracted per image, and two different COLMAP feature matching algorithms: sequential matching with loop closures and Vocabulary Tree Matcher. 

The results are summarised in \tabref{tab:pose_ablation}. For all COLMAP configurations, providing the SLAM prior poses not only accelerates pose computation, but also leads to better test rendering, compared to the offline COLMAP. Our SLAM prior poses also register all images in the trajectory, while when not provided, COLMAP only registers 55\%-95\% images. Extracting more visual features per image (from 1024 to 8192) leads to higher percentage of image registration and better visual reconstruction (PSNR and SSIM). This comes at the expense of a higher computation time, especially with the VocabTree matcher. Using the COLMAP Sequential Matcher is generally faster than Vocabulary Tree Matcher.

\subsection{Submapping for Large-Scale Environments}

\begin{figure}[t]
	\centering
	\includegraphics[width=0.80\columnwidth]{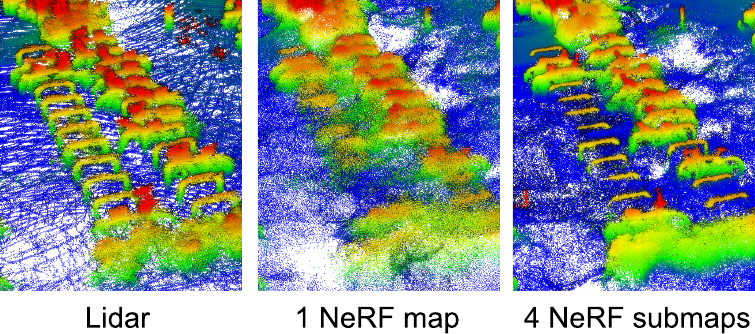}
	\caption{Effect of submap size on reconstruction quality. A larger number of submaps for a given area results in better reconstruction on thin objects such as the bike racks on the right.}
 \vspace{-8pt}
	\label{fig:submap_recon}
\end{figure}

We show a large-scale reconstruction of HBAC-Maths with the handheld Frontier using submapping in \figref{fig:header}, as well as the trajectory for each submap. To demonstrate the advantage of submapping, we compare the 3D reconstruction and rendering quality using one NeRF model for the entire sequence versus NeRF models built with multiple submaps. We present the qualitative results in \figref{fig:submap_recon}. The reconstruction of the bike racks in Maths Inst. is blurred when only using a single NeRF map due to its limited representation capability for storing all objects over a large area. While using only a dedicated submap for that local area, the reconstruction quality improves significantly as seen in \figref{fig:submap_recon}~(right). 

\section{CONCLUSIONS}
In summary, we proposed a large-scale 3D reconstruction system fusing both lidar and vision in a neural field via differentiable rendering. The proposed sensor fusion overcomes the limitations of individual sensors, namely the sparsity of the lidar and the fragility of vision in the presence of texture-less surface, and limited multi-view constraints. We demonstrate large-scale reconstruction results from real-world datasets collected from multiple robot platforms in conditions suited for inspection tasks.

\section*{Acknowledgment}
The authors would like to thank Ren Komatsu for his help on software development, Tobit Flatscher for deploying spot robot, Rowan Border for drone data collection, and Sundara Tejaswi Digumarti for his helpful discussions.


\bibliographystyle{IEEEtran}
\bibliography{IEEEabrv, references}

\end{document}